\documentclass[runningheads]{llncs}
\usepackage[T1]{fontenc}
\usepackage{lmodern}
\usepackage{amsmath,amssymb,graphicx,booktabs}
\usepackage{cite}
\usepackage{placeins}
\usepackage[hidelinks]{hyperref}
\newcommand{\R}{\mathcal{R}}
\newcommand{\A}{\mathcal{A}}
\newcommand{\Iop}{\mathcal{I}}
\newcommand{\M}{\mathcal{M}}
\newcommand{\Th}{\Theta}
\newcommand{\x}{\mathbf{x}}
\newcommand{\pp}{\mathbf{p}}
\newcommand{\muw}{\mu_{\mathrm{w}}}
\newcommand{\lmax}{\lambda_{\max}}
\newcommand{\dc}{\textsc{DC-DuNet}}
\newcommand{\imgonly}{\textsc{Image-UNet}}
\newcommand{\duald}{\textsc{Dual-domain}}
\newcommand{\sinoinit}{\dc\ (sinogram init.)}
\newcommand{\interpfbp}{\textsc{Interp-FBP}}
\newcommand{\sparsefbp}{\textsc{FBP}}

\newcommand{\doseRef}{10\%}
\newcommand{\numViewsRef}{60}
\newcommand{\paperBaseImg}{16}
\newcommand{\paperCompErrDense}{0.0004}
\newcommand{\paperCompErrMid}{0.0011}
\newcommand{\paperCompErrSparse}{0.0035}
\newcommand{\paperCompRatioHi}{2563}
\newcommand{\paperCompRatioLo}{286}
\newcommand{\paperCompletionPct}{99.7}
\newcommand{\paperConventionDelta}{0.05}
\newcommand{\paperDepth}{3}
\newcommand{\paperEpochs}{20}
\newcommand{\paperExtDelta}{-1.14}
\newcommand{\paperExtSinoDelta}{1.28}
\newcommand{\paperExtSinoSlices}{13}
\newcommand{\paperExtSinoWins}{13}
\newcommand{\paperExtSlices}{13}
\newcommand{\paperExtWins}{0}
\newcommand{\paperGainAblInteraction}{0.16}
\newcommand{\paperGainCompletionAC}{1.27}
\newcommand{\paperGainCond}{-0.62}
\newcommand{\paperGainDCC}{0.33}
\newcommand{\paperGainImageBase}{-2.16}
\newcommand{\paperGainImageBaseOnInit}{-2.32}
\newcommand{\paperGainSinoInit}{1.98}
\newcommand{\paperGainSinoInitOnBase}{1.82}
\newcommand{\paperInterpDb}{4.65}
\newcommand{\paperLambdaBandHi}{2}
\newcommand{\paperLambdaBandLo}{0.1}
\newcommand{\paperLambdaBandSpread}{0.47}
\newcommand{\paperLambdaCliffAfter}{34.99}
\newcommand{\paperLambdaCliffAt}{3}
\newcommand{\paperLambdaCliffBefore}{37.20}
\newcommand{\paperLambdaCliffDelta}{2.21}
\newcommand{\paperLambdaCliffGrid}{256}
\newcommand{\paperLambdaCoef}{0.85}
\newcommand{\paperLambdaMax}{2.0}
\newcommand{\paperLambdaNRange}{64--512}
\newcommand{\paperLambdaR}{0.9999}
\newcommand{\paperLambdaResid}{0.90}
\newcommand{\paperLambdaSpread}{1.08}
\newcommand{\paperLambdaStable}{2.35}
\newcommand{\paperN}{256}
\newcommand{\paperNTest}{40}
\newcommand{\paperNTrain}{180}
\newcommand{\paperNVal}{30}
\newcommand{\paperParamPct}{33}
\newcommand{\paperParamRatio}{1.33}
\newcommand{\paperParamsImage}{0.515}
\newcommand{\paperParamsOurs}{0.686}
\newcommand{\paperRealExtDelta}{-0.11}
\newcommand{\paperRealExtPatients}{10}
\newcommand{\paperRealExtQPatient}{0.093}
\newcommand{\paperRealExtSinoDelta}{0.12}
\newcommand{\paperRealExtSinoQ}{1.5\times10^{-3}}
\newcommand{\paperRealExtSinoQPatient}{5.1\times10^{-3}}
\newcommand{\paperRealExtSinoSlices}{13}
\newcommand{\paperRealExtSinoWins}{13}
\newcommand{\paperRealExtSlices}{13}
\newcommand{\paperRealExtWins}{3}
\newcommand{\paperRealFullVsImageRef}{0.02}
\newcommand{\paperRealImageBase}{0.00}
\newcommand{\paperRealPatients}{10}
\newcommand{\paperRealSigMinQPatient}{0.065}
\newcommand{\paperRealSigPairs}{240}
\newcommand{\paperRealSigWins}{183}
\newcommand{\paperRealSigWorstQ}{0.081}
\newcommand{\paperRealSinoInit}{0.20}
\newcommand{\paperRealSinoVsImgDense}{0.04}
\newcommand{\paperRealSinoVsImgSparse}{0.10}
\newcommand{\paperRealSlices}{40}
\newcommand{\paperSigMaxDelta}{1.75}
\newcommand{\paperSigMinDelta}{0.97}
\newcommand{\paperSigPairs}{240}
\newcommand{\paperSigRefDelta}{1.36}
\newcommand{\paperSigSinoMinDelta}{-0.99}
\newcommand{\paperSigSinoRefDelta}{0.96}
\newcommand{\paperSigSinoRefN}{40}
\newcommand{\paperSigSinoRefQ}{4.3\times10^{-8}}
\newcommand{\paperSigSinoRefWins}{40}
\newcommand{\paperSigWins}{240}
\newcommand{\paperSigWorstQ}{3.6\times10^{-8}}
\newcommand{\paperVfull}{720}
\newcommand{\paperVinterp}{180}

\begin{document}
\title{Angular Interpolation in Sparse-View Low-Dose CT Reconstruction: Placement Effects and Data-Consistency Step-Size Scaling}
\titlerunning{Angular Interpolation for Sparse-View Low-Dose CT}
\author{Yitian Tang\inst{1}\thanks{Yitian Tang and Yuncheng Zhang contributed equally to this work.} \and
Yuncheng Zhang\inst{1}\textsuperscript{\ensuremath{\star}} \and
Hualiang Wang\inst{2} \and
Zuozhu Liu\inst{1}\thanks{Corresponding author.}}
\authorrunning{Y. Tang et al.}
\institute{Zhejiang University--University of Illinois Urbana-Champaign Institute,\\
Zhejiang University, Hangzhou, China\\
\email{\{yitian.24,yuncheng.24,zuozhuliu\}@intl.zju.edu.cn}
\and
Zhejiang University, Hangzhou, China\\
\email{hualiang\_wang@zju.edu.cn}}
\maketitle

\begin{abstract}
Dual-domain networks combine projection and image processing for sparse-view low-dose computed tomography (CT). We quantify the contribution of angular interpolation before learning and distinguish its roles in sinogram initialisation and the image residual base. On synthetic parallel-beam data, parameter-free interpolation reduces noise-free missing-view mean absolute error by \paperCompletionPct\% relative to zero filling at \numViewsRef\ views. Full-view filtered back-projection (FBP) of the jointly learned completed sinogram adds \paperGainCompletionAC\,dB of peak signal-to-noise ratio (PSNR) over interpolated FBP. In a data-consistency dual-domain network (\dc), a $2\times2$ configuration ablation controls branch width, depth and training schedule while varying these placements and their associated input channels. After \paperEpochs\ epochs, the PSNR contrasts are $+\paperGainSinoInit$\,dB for sinogram initialisation and $\paperGainImageBase$\,dB for the interpolated image base; the image-base ranking also reverses during training. The sinogram-initialised variant exceeds an image-only U-Net by \paperSigSinoRefDelta\,dB at 60 views and 10\% dose, and by \paperExtSinoDelta\,dB on independently generated synthetic data. The U-Net leads at higher view counts and with a shared interpolated image base. Without retraining, the variant retains a \paperRealExtSinoDelta\,dB advantage on held-out CHAOS abdominal CT under simulated degradation; the image-base contrast rounds to zero. The implemented projection/\allowbreak backprojection pair has empirical dominant gain $\approx\paperLambdaCoef\,NK$, where $N$ is image-grid width and $K$ acquired view count. These results distinguish interpolation placement from network architecture and support $NK$ normalisation for step calibration.

\end{abstract}

\section{Introduction}
\label{sec:intro}

Computed tomography contributes substantially to medical radiation exposure~\cite{brenner}. Reducing projection views and photon counts offers two routes to dose reduction, with distinct reconstruction challenges: missing angles produce streak artefacts, while lower photon counts introduce measurement noise. Sparse-view low-dose reconstruction must address both.

Image-domain methods restore reconstructed images. RED-CNN~\cite{redcnn} uses a residual encoder--decoder for low-dose denoising. FreeSeed~\cite{freeseed} combines frequency-band-aware processing with self-guided refinement for sparse-view artefacts, while CoreDiff~\cite{corediff} uses contextual error-modulated diffusion for low-dose denoising and generalisation. These methods learn image priors, but the input reconstruction determines which artefacts and residual errors they must process.

Projection information constrains reconstruction. DRONE~\cite{drone} combines data- and image-domain refinement; learned primal--dual reconstruction~\cite{learnprimaldual} embeds forward and adjoint operators in an iterative network. MoDL~\cite{moDL} combines a learned regulariser with model-based data consistency. Score-based generative modelling~\cite{songscore} supplies a probabilistic prior, while CvG-Diff~\cite{cvgdiff2025} treats sparse-view reconstruction as cross-view generalised diffusion. Primitive-based representations also address extremely sparse-view cone-beam CT~\cite{lingauss2024}.

Angular interpolation estimates missing parallel-beam projections without learned parameters, using the angular structure of the Radon transform~\cite{kakslaney}. It can enter a dual-domain network as the sinogram initialisation or, after filtered back-projection (FBP), as the image residual base. Separating these placements distinguishes analytic preprocessing from the learned architecture.

At the reference setting, interpolation followed by FBP improves PSNR over sparse-view FBP by \paperInterpDb\,dB without learned parameters. Full-view FBP of the jointly learned completed sinogram adds \paperGainCompletionAC\,dB over interpolated FBP. With a shared interpolated image base, a smaller image-only U-Net achieves higher PSNR than the full data-consistency dual-domain network (\dc). We therefore evaluate interpolation placement separately from network architecture.

First, a $2\times2$ configuration ablation compares sinogram initialisation and the image residual base, including associated inputs, at matched branch width, depth and training schedule. After 20 epochs, the two PSNR contrasts have opposite signs, and the image-base ranking reverses during training (\S\ref{sec:ablation}). The sinogram-initialised variant leads at the sparsest tested operating points, with a ranking crossover as view count increases (\S\ref{sec:crossover}). CHAOS evaluation tests transfer to real anatomy under simulated degradation (\S\ref{sec:real}).

Second, we measure the grid dependence of the data-consistency step. The implemented projection/backprojection pair has dominant gain $\approx\paperLambdaCoef\,NK$, exposing the grid-width factor omitted by view-count-only normalisation. We derive scaling under exact-adjoint conditions and evaluate $NK$ normalisation across image grids (\S\ref{sec:dcc}, \S\ref{sec:dcc-analysis}). This operator analysis calibrates consistency steps in the tested 2-D parallel-beam setting.

\section{Method}
\label{sec:method}

\begin{figure}[t]
  \centering
  \includegraphics[width=\textwidth]{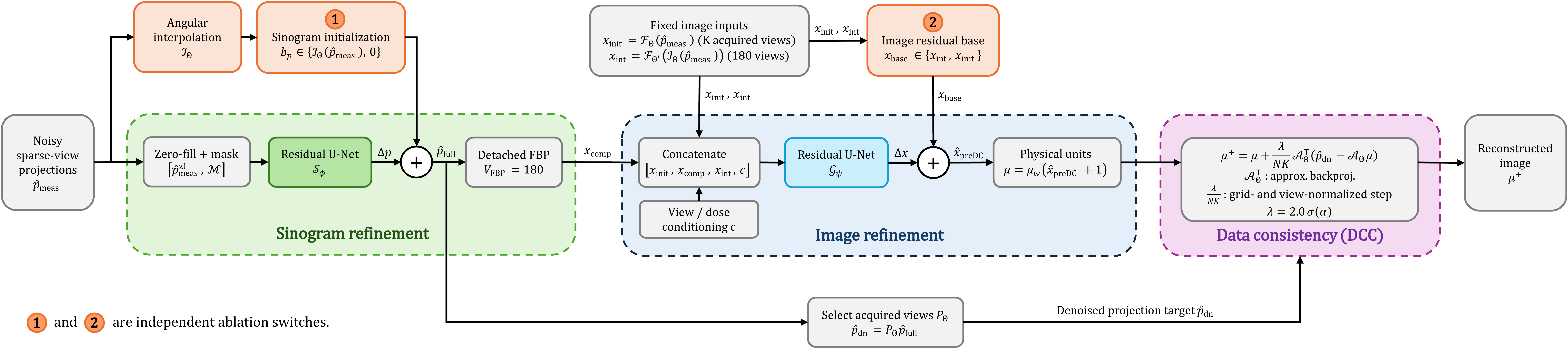}
  \caption{The pipeline. Sinogram refinement, detached FBP and image residual refinement precede DCC in attenuation units, using denoised acquired-view projections from the completed sinogram. Independent switches control sinogram initialisation (\textcircled{1}) and the image residual base (\textcircled{2}); associated inputs also change in the ablation. The adjoint symbol denotes the approximate backprojector (\S\ref{sec:dcc}).}

  \label{fig:overview}
\end{figure}

Figure~\ref{fig:overview} shows the interpolation-first pipeline and data-consistency correction (DCC).

\subsection{Problem setup}
\label{sec:setup}

Write $\mu$ for the linear attenuation coefficient of a slice in
$\mathrm{cm}^{-1}$ on an $N\times N$ grid, and $\R_\Th$ for the discretised Radon
transform on a parallel-beam geometry with angle set
$\Th=\{\theta_1,\dots,\theta_K\}$.  A full scan takes $V_{\text{full}}=\paperVfull$
equidistant angles over $180^\circ$; a sparse scan takes $K\ll V_{\text{full}}$ of
them, recorded by the mask $\M\in\{0,1\}^{V_{\text{full}}}$.  Two degradations are applied to the clean sinogram $\pp$.  Sparse-view sampling keeps $\M\odot\pp$.
The acquired rows undergo Poisson photon counting at a dose-dependent reference intensity followed by a log transform. Electronic read noise and detector blur are disabled in the reported experiments. At fixed attenuation, the noise scale is approximately $\sigma\propto\text{dose}^{-1/2}$.  Regression targets the network-unit image
$\x^\star=\mathcal{F}_{\text{full}}(\pp)$, where $\mathcal{F}_\Th$ denotes FBP on
$\Th$ and network units are the affine rescaling $\x=(\mu-\muw)/\muw$ with
$\muw=0.192\,\mathrm{cm}^{-1}$.  This rescaling maps water to $0$ and air to $\approx-1$, allowing a shared learning rate across prediction heads.  Evaluation compares reconstructions with the original Hounsfield-unit (HU) images. PSNR and the structural similarity index (SSIM)~\cite{ssim} use images clipped and rescaled within the $W/L=3000/500$~HU window; mean absolute error (MAE) uses unwindowed HU values.

\subsection{Interpolation-first parameterisation}
\label{sec:interp-first}

Nearby views in a parallel-beam sinogram have related projection profiles~\cite{kakslaney}, allowing angular interpolation to estimate missing rows of $\M\odot\pp$. Let $\Iop_\Th$ linearly interpolate between bracketing acquired views, with nearest-view extension at the angular boundaries. The interpolant reproduces acquired views \emph{exactly}; the learned residual may subsequently modify them.  With noise disabled and zero-fill missing-view MAE normalised to $1.0$, linear interpolation reduces this error against the clean full sinogram to $\paperCompErrSparse$ / $\paperCompErrMid$ / $\paperCompErrDense$ at $K=60$ / $120$ / $240$. The reduction factor is $\paperCompRatioLo$--$\paperCompRatioHi$ and increases with $K$.

Completion combines closed-form initialisation and a learned residual:
\begin{equation}
\label{eq:completion}
\hat\pp_{\text{full}}
= \underbrace{\Iop_\Th\big(\hat\pp_{\text{meas}}\big)}_{\text{closed form}}
\;+\; \underbrace{\mathcal{S}_\phi\big(\hat\pp_{\text{meas}}, \M\big)}_{\text{learned residual}},
\end{equation}
where $\mathcal{S}_\phi$ is a sinogram U-Net~\cite{unet}. The interpolant estimates the missing views; $\mathcal{S}_\phi$ is trained to suppress noise on \emph{acquired} views and refine the completed sinogram for consistency with the denoised projections, reducing propagation of interpolation error into the image domain.

The image branch also predicts a residual. We distinguish the acquired-view FBP base from the interpolated-FBP base:
\begin{equation}
\label{eq:bases}
\x_{\text{init}} = \mathcal{F}_\Th\big(\hat\pp_{\text{meas}}\big),
\qquad
\x_{\text{int}} = \mathcal{F}_{\Th'}\big(\Iop_\Th(\hat\pp_{\text{meas}})\big),
\end{equation}
where $\Th'$ retains a fixed $V_{\text{interp}}=\paperVinterp$ views to limit the cost of NumPy FBP.  Both bases require no learned parameters, but they are not equivalent.
$\x_{\text{init}}$ contains stronger streak artefacts; $\x_{\text{int}}$ reduces
them but retains noise and interpolation error.  The full \dc, \duald\ and image-only baseline use $\hat\x=\x_{\text{int}}+\mathcal{G}_\psi(\cdot)$ in the primary architecture comparison, holding the residual base fixed. We separately compare interpolation placements in \S\ref{sec:ablation}.

\subsection{Network architecture and shared controls}
\label{sec:arch}

The full \dc\ combines sinogram refinement, image residual refinement, conditioning and DCC; \S\ref{sec:ablation} compares their configurations. Its \emph{sinogram branch} $\mathcal{S}_\phi$ maps the zero-filled measurement and mask to a residual on the interpolated sinogram.  It operates at half angular resolution and upsamples the \emph{residual}, rather than the sinogram. The interpolation base retains the acquired measurements.  The \emph{image branch} $\mathcal{G}_\psi$
predicts the residual of the parameterisation from
$[\x_{\text{init}},\x_{\text{comp}},\x_{\text{int}},c]$ (sparse FBP, completed-sinogram FBP, interpolated base and conditioning maps). \imgonly\ uses this branch alone, omitting $\x_{\text{comp}}$ and $c$. The \duald\ baseline denoises acquired-view projections and supplies their detached FBP reconstruction to an image residual branch with the $\x_{\text{int}}$ base; it omits conditioning and DCC.  \emph{Conditioning}
adds $c=[K/V_{\text{full}},\mathrm{clip}((\log_{10}\text{dose}+2)/2,-1,2)]$, as
feature-wise linear modulation (FiLM)~\cite{film} at the bottleneck and as constant input channels, so that one model serves
every view/dose setting.  The output heads and final FiLM layers are zero-initialised, so the initial residual predictions are zero. The DCC step can
subsequently modify the image base.  Training minimises a weighted sum of a Charbonnier~\cite{charbonnier} image term, a multiscale difference-of-Gaussians band loss, gradient and Laplacian terms, supervision against the clean full sinogram and clean \emph{acquired} projections, and a Radon consistency term on $48$ views. Applicable losses use common weights across models.
The NumPy FBP feature computation is detached from the gradient path.  The measured-views term is supervised with the clean
acquired projections, not the noisy measurement.

\subsection{Data-consistency update and step normalisation}
\label{sec:dcc}

A Radon consistency loss supervises projection agreement during training. The network also applies a differentiable data-consistency update during both training and inference,
\begin{equation}
\label{eq:dcc}
\mu \leftarrow \mu + \frac{\lambda}{NK}\,
B_\Th\big(\hat\pp_{\text{dn}} - \A_\Th\mu\big),
\qquad \lambda = \lmax\,\sigma(\theta),
\end{equation}
with $\A_\Th$ a differentiable rotate-and-sum Radon transform and $\sigma$
the logistic function. Here $B_\Th$ is the implemented approximate
backprojector, and $\lmax$ without an operator argument denotes the
step-size cap rather than a measured eigenvalue. The updated attenuation
is mapped back to network units as $\x=(\mu-\muw)/\muw$.  Three implementation choices affect this update. (i)~Compute the projection residual in attenuation units. The projection is linear in $\mu$ but affine in $\x=(\mu-\muw)/\muw$, requiring conversion before projection and after the update. (ii)~Use the \emph{denoised} sinogram as the target: setting
$\hat\pp_{\text{dn}}=\hat\pp_{\text{meas}}$ reintroduces Poisson noise into the image and counteracts denoising. (iii)~Normalise by both $N$ and $K$. For two iterates with the same fixed
projection target, their difference evolves as
$\delta\x\leftarrow(I-\tfrac{\lambda}{NK}B_\Th A_\Th)\delta\x$;
the $\muw$ factors from the unit conversions cancel. In the ideal case
$B_\Th=A_\Th^{\!\top}$, a positive step $\lambda_{\mathrm{abs}}=\lambda/(NK)$ satisfying
$\lambda_{\mathrm{abs}}\lmax(A_\Th^{\!\top}A_\Th)<2$ contracts modes with
positive eigenvalues, while null-space modes remain unchanged. Power iteration on the
implemented $B_\Th A_\Th$ gives a dominant gain approximately proportional to
$\paperLambdaCoef\,NK$ over $N\in\paperLambdaNRange$ and
$K\in\{30,60,120,240,720\}$, motivating $NK$ scaling. For the approximate backprojector, we bound the learned step with $\lmax=\paperLambdaMax$ and assess finite-rollout stability empirically.

\section{Experiments}
\label{sec:exp}

\subsection{Data, degradation and training schedule}
\label{sec:impl}

Training and in-distribution evaluation use synthetic anthropomorphic abdominal
slices from a procedural generator of randomised ellipse phantoms. Dose reduction is simulated by lowering incident photon fluence, an approach also used in low-dose CT benchmarks~\cite{moen2021}.
The grid
is $N=\paperN$, a $400$\,mm field of view and $V_{\text{full}}=\paperVfull$ views;
the split is \paperNTrain\ training slices, \paperNVal\ validation, and
\paperNTest\ test slices never seen in training.  View count and dose are drawn
per training batch. Evaluation fixes each view/dose cell and averages over
slices with reproducibly seeded noise draws.  A second held-out set of independently generated synthetic slices tests transfer within the procedural generator, and \S\ref{sec:real}
adds real human abdominal CT from the CHAOS challenge~\cite{chaos}.  Every learned model uses a 20-epoch budget and batch size 4. Dual-domain models use 2 sinogram warm-up epochs followed by 18 joint epochs; \imgonly\ trains its image branch for all 20 epochs. Models share branch width and depth (\paperBaseImg\ image base channels,
8 sinogram base channels where applicable, depth \paperDepth) and common weights for applicable losses. The interpolation ablation switches
also remove the corresponding $\x_{\text{comp}}$ or $\x_{\text{int}}$ input
channel, so their contrasts include changes in feature availability.  \dc\ carries \paperParamsOurs\,M parameters against \paperParamsImage\,M
for \imgonly\ (\paperParamPct\% more).  Evaluation noise is seeded per batch ($\text{seed}+\text{batch
index}$), ensuring that paired comparisons use the same noise realisations.  Inference uses one forward pass per domain and the iterations of \eqref{eq:dcc}.

\subsection{Interpolation placement ablation}
\label{sec:ablation}

\begin{table}[t]
  \caption{Ablation at $K=\numViewsRef$, dose $=\doseRef$. Top four rows vary sinogram initialisation and image residual base, including associated input channels, at fixed width/depth and schedule. $\Delta$ is relative to full \dc\ (first row); bottom rows disable conditioning or DCC. Bold: best PSNR/SSIM.}

  \label{tab:ablation}
  \setlength{\tabcolsep}{3pt}
  \begin{tabular*}{\columnwidth}{@{\extracolsep{\fill}}llrrr@{}}
    \toprule
    \shortstack[l]{Sinogram\\initialisation} & Image base & PSNR [dB] & SSIM & $\Delta$ [dB]\\
    \midrule
    Interpolated & $\x_{\text{int}}$ & 41.91 & 0.9824 & --\\
    Zero & $\x_{\text{int}}$ & 40.10 & 0.9703 & $-1.82$\\
    Interpolated & $\x_{\text{init}}$ & \textbf{44.23} & 0.9800 & $+2.32$\\
    Zero & $\x_{\text{init}}$ & 42.26 & 0.9609 & $+0.35$\\
    \midrule
    \multicolumn{2}{@{}l}{\shortstack[l]{Without view/dose\\conditioning}} & 42.53 & \textbf{0.9883} & $+0.62$\\
    \multicolumn{2}{@{}l}{\shortstack[l]{Without data\\consistency}} & 41.58 & 0.9789 & $-0.33$\\
    \bottomrule
  \end{tabular*}
\end{table}

Table~\ref{tab:ablation} varies sinogram initialisation and image residual base at fixed width/\allowbreak depth, schedule and applicable loss weights. Relative to zero sinogram initialisation with acquired-view FBP base, interpolated sinogram initialisation changes PSNR by $+\paperGainSinoInit$\,dB and the interpolated image base by $\paperGainImageBase$\,dB. With the other switch enabled, the respective contrasts are $+\paperGainSinoInitOnBase$ and $\paperGainImageBaseOnInit$\,dB; their signed interaction is $-\paperGainAblInteraction$\,dB. The highest-PSNR configuration, \sinoinit, retains interpolated sinogram initialisation but uses acquired-view FBP as its image base. We use this name consistently to distinguish the variant from full \dc, which uses the interpolated-FBP image base.

The interpolated image base initially gives higher PSNR, but the training curves cross within nine epochs. The ranking therefore depends on training duration. Retaining conditioning changes reference-cell PSNR by \paperGainCond\,dB; retaining DCC changes it by $+\paperGainDCC$\,dB. All configurations use the same fixed training seed.

\begin{figure}[t]
  \centering
  \includegraphics[width=\textwidth]{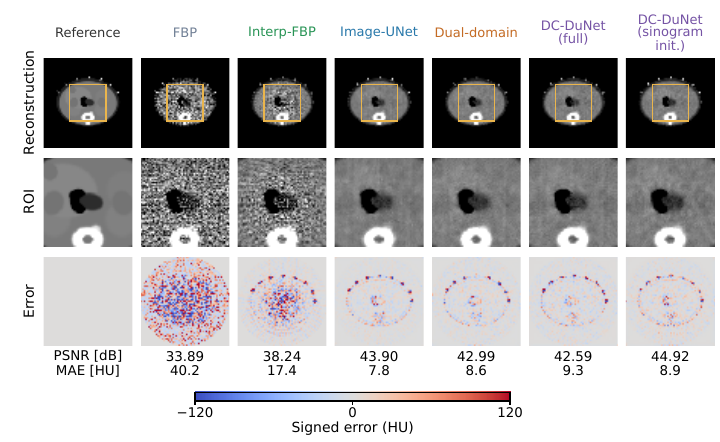}
  \caption{Reconstructions of the same synthetic test slice at $K=\numViewsRef$, dose $=\doseRef$. Columns: reference, FBP, Interp-FBP, Image-UNet, Dual-domain, full \dc\ and the sinogram-initialised variant. Rows: reconstruction, boxed ROI and signed error (prediction minus reference). Display $W/L=400/50$~HU; error scale $\pm120$~HU. Values below the error maps are displayed-slice PSNR ($W/L=3000/500$~HU) and MAE (HU). Dataset means are reported in Table~\ref{tab:main}.}

  \label{fig:qual}
\end{figure}

\subsection{Architecture comparisons across view counts}
\label{sec:crossover}

\begin{table}[t]
  \caption{Controlled architecture comparison: PSNR [dB] on identical degraded measurements ($W/L=3000/500$~HU); SSIM at $K=\numViewsRef$, dose $=\doseRef$. Oracle: clean full-view FBP. Sinogram init.: interpolated sinogram with acquired-view FBP base. Bold: column best excluding oracle.}

  \label{tab:main}
  \setlength{\tabcolsep}{3pt}
  \begin{tabular*}{\columnwidth}{@{\extracolsep{\fill}}lcccc@{}}
    \toprule
    method & $K{=}60$ & $K{=}60$ & $K{=}120$ & SSIM\\
    & $10$\% & $5$\% & $10$\% & (ref.)\\
    \midrule
    \sparsefbp   & 32.70 & 30.05 & 36.39 & 0.6352\\
    \interpfbp   & 37.35 & 35.62 & 39.87 & 0.9189\\
    \imgonly     & 43.27 & 42.41 & \textbf{47.20} & \textbf{0.9879}\\
    \duald       & 42.49 & 41.85 & 46.88 & 0.9862\\
    \dc & 41.91 & 41.29 & 46.23 & 0.9824\\
    \emph{oracle} & 52.50 & 52.50 & 52.50 & 0.9991\\
    \midrule
    \shortstack[l]{\dc\\(sinogram init.)} & \textbf{44.23} & \textbf{42.74} & 46.62 & 0.9800\\
    \bottomrule
  \end{tabular*}
\end{table}

Table~\ref{tab:main} reports three settings from the controlled architecture comparison; paired tests cover six fixed operating points. \imgonly\ and \duald\ are the image-only and dual-domain baselines defined in \S\ref{sec:arch}. Models receive identical slices and noise realisations. Confidence intervals use a slice-level cluster bootstrap. Two-sided Wilcoxon signed-rank tests give paired $p$ values, with Benjamini--Hochberg adjustment across operating points yielding $q$ values. With the shared interpolated-FBP base, \imgonly\ exceeds full \dc\ by \paperSigMinDelta--\paperSigMaxDelta\,dB (\paperSigWins/\paperSigPairs\ slice comparisons, worst $q=\paperSigWorstQ$), using approximately $1/\paperParamRatio$ of the parameters.

The sinogram-initialised configuration exceeds \imgonly\ by \paperSigSinoRefDelta\,dB at $K=60$, dose $=10\%$ (\paperSigSinoRefWins/\paperSigSinoRefN\ slices, $q=\paperSigSinoRefQ$). It leads at the two sparsest settings; \imgonly\ leads from 120 views onward, with a $|\paperSigSinoMinDelta|$\,dB margin at 240 views. The reference-cell results show a metric trade-off: the sinogram-initialised variant has the highest PSNR, while \imgonly\ has the highest SSIM. Figure~\ref{fig:qual} compares reconstruction and error patterns on the same measurements for these residual-base and input configurations.

\subsection{Real anatomy}
\label{sec:real}

\begin{table}[t]
  \caption{CHAOS anatomy under simulated degradation, without retraining. PSNR deltas are A$-$B [dB]; $q$ is dimensionless. PSNR uses the $W/L=3000/500$~HU quality window. Parentheses: wins/total slices; view-count comparisons use dose $=10\%$.}

  \label{tab:real}
  \small
  \setlength{\tabcolsep}{4pt}
  \begin{tabular*}{\columnwidth}{@{\extracolsep{\fill}}lcc@{}}
    \toprule
    comparison / statistic & synthetic & real\\
    \midrule
    \imgonly\ $-$ \dc\ (ref.\ point) & $+\paperSigRefDelta$ &
      $-\paperRealFullVsImageRef$\\
    \quad worst adjusted $q$ & $\paperSigWorstQ$ & $\paperRealSigWorstQ$\\
    \midrule
    Sinogram initialisation effect ($2{\times}2$) & $+\paperGainSinoInit$ &
      $+\paperRealSinoInit$\\
    Interpolated-FBP base effect ($2{\times}2$) & $\paperGainImageBase$ &
      $+\paperRealImageBase$\\
    \midrule
    \shortstack[l]{\sinoinit\\$-$ \imgonly\ ($K{=}60$)} & $+\paperSigSinoRefDelta$ &
      $+\paperRealSinoVsImgSparse$\\
    \shortstack[l]{\sinoinit\\$-$ \imgonly\ ($K{=}240$)} & $\paperSigSinoMinDelta$ &
      $+\paperRealSinoVsImgDense$\\
    \shortstack[l]{\sinoinit\\$-$ \imgonly\ (held-out)} & $+\paperExtSinoDelta$
      $(\paperExtSinoWins/\paperExtSinoSlices)$ & $+\paperRealExtSinoDelta$
      $(\paperRealExtSinoWins/\paperRealExtSinoSlices)$\\
    \shortstack[l]{\dc\ (full)\\$-$ \imgonly\ (held-out)} & $\paperExtDelta$
      $(\paperExtWins/\paperExtSlices)$ & $\paperRealExtDelta$
      $(\paperRealExtWins/\paperRealExtSlices)$\\
    \bottomrule
  \end{tabular*}
\end{table}

\begin{figure}[p]
  \centering
  \includegraphics[width=\textwidth]{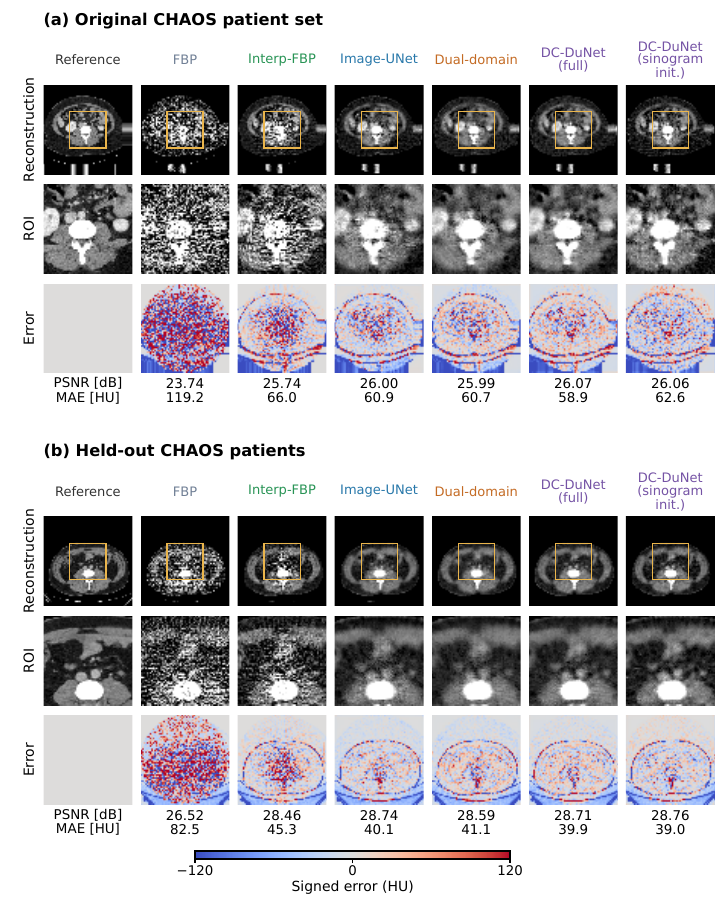}
  \caption{CHAOS examples under simulated degradation at $K=60$, dose $=10\%$, without retraining: (a) the original patient group and (b) held-out patients. Each displayed slice is the first reference slice saved by its evaluation. Column and row ordering, ROI marking and metric placement follow Figure~\ref{fig:qual}. Display $W/L=400/50$~HU; both groups share the $\pm120$~HU signed-error scale. PSNR ($W/L=3000/500$~HU) and MAE (HU) are for each displayed slice; aggregate comparisons remain in Table~\ref{tab:real}.}

  \label{fig:real-qual}
\end{figure}

We evaluate \paperRealSlices\ CHAOS slices from \paperRealPatients\ patients and \paperRealExtSlices\ held-out slices from \paperRealExtPatients\ other patients, resampled, projected and degraded by the same pipeline. The existing trained models are applied without retraining. Table~\ref{tab:real} quantifies transfer through simulated projection and degradation, and Figure~\ref{fig:real-qual} adds reconstruction, local-detail and error views.

The sinogram-initialised configuration exceeds \imgonly\ by \paperRealExtSinoDelta\,dB on held-out patients (\paperRealExtSinoWins/\paperRealExtSinoSlices\ slice wins, $q=\paperRealExtSinoQ$). The held-out advantage has the same direction on real and synthetic anatomy, with gains of 0.12 and 1.28\,dB, respectively. The synthetic image-base contrast of \paperGainImageBase\,dB rounds to 0.00\,dB on real anatomy, showing a different image-base contrast between procedural phantoms and CHAOS anatomy.

Full \dc\ leads \imgonly\ in all six real-anatomy in-distribution cells (\paperRealSigWins/\paperRealSigPairs\ slice comparisons, worst $q=\paperRealSigWorstQ$), reversing the synthetic ranking. The corresponding patient-level tests have $q\geq\paperRealSigMinQPatient$. On held-out real patients, \imgonly\ has higher mean PSNR than full \dc\ (\paperRealExtWins/\paperRealExtSlices\ full-model wins, patient-level $q=\paperRealExtQPatient$). Thus, the full-model ranking changes between procedural phantoms, the original CHAOS group and held-out patients.

\FloatBarrier

\subsection{The data-consistency step}
\label{sec:dcc-analysis}

Power iteration gives a dominant gain of approximately $\paperLambdaCoef\,NK$ ($r=\paperLambdaR$); the maximum fit residual relative to the largest measured gain is \paperLambdaResid\%. Gain divided by $NK$ varies by only \paperLambdaSpread$\times$, whereas gain divided by $K$ scales with $N$. Across $N=64,128,256$, eight-step training-free rollouts have a maximum within-grid PSNR spread of \paperLambdaBandSpread\,dB for $\lambda\in[\paperLambdaBandLo,\paperLambdaBandHi]$. At $N=\paperLambdaCliffGrid$, increasing $\lambda$ from $\paperLambdaBandHi$ to $\paperLambdaCliffAt$ lowers PSNR from \paperLambdaCliffBefore\ to \paperLambdaCliffAfter\,dB, a \paperLambdaCliffDelta\,dB drop across the approximate threshold $2/\paperLambdaCoef\approx\paperLambdaStable$. An extra division of backprojection by $\muw$ inflates the gain by about $5.2\times$ and yields an incorrect $\lambda<0.45$ bound. With steps calibrated at $N=64$, the $NK$- and $K$-normalised conventions differ by at most approximately \paperConventionDelta\,dB over the tested grids. $NK$ normalisation therefore calibrates steps across the tested grid sizes, with the empirical threshold evaluated by finite rollouts for this discretisation and approximate backprojector.

\section{Discussion}
\label{sec:discussion}

The placement ablation (Table~\ref{tab:ablation}) distinguishes analytic sinogram initialisation from the image residual base. The sinogram-initialised variant has the highest reference-cell PSNR and leads at the sparsest synthetic view counts. With a shared interpolated image base, the smaller \imgonly\ leads full \dc. These configurations include associated input channels; the training-curve crossover also makes the schedule relevant to their comparison. The PSNR/SSIM trade-off distinguishes reconstruction fidelity from structural similarity.

On held-out real anatomy, patient-level resampling supports the sinogram-initialised advantage ($q=\paperRealExtSinoQPatient$). In the original CHAOS group, the best patient-level value is $q=\paperRealSigMinQPatient$, and the image-base contrast rounds to zero. The gain transfers under simulated degradation, while its magnitude and the image-base contrast differ between procedural phantoms and CHAOS anatomy.

The evaluation uses 2-D parallel-beam geometry, procedural training phantoms, simulated photon noise and one training seed per configuration. Rankings correspond to the \paperEpochs-epoch schedule; repeated seeds and longer schedules would test their persistence. Evaluation with acquired projections and cone-beam or helical geometry also requires consistent peripheral padding and field of view (FOV).

\section{Conclusion}
\label{sec:conclusion}

Angular interpolation yields opposite PSNR contrasts in sinogram-initialisation and image-residual-base configurations after 20 epochs. The training-curve crossover shows that measured gains depend on configuration and training duration. The sinogram-initialised variant leads at the sparsest synthetic view counts and on held-out real anatomy under simulated degradation. The empirical $\paperLambdaCoef\,NK$ operator gain calibrates data-consistency steps across the tested image grids.

\end{document}